\documentclass{article} % For LaTeX2e
\usepackage{iclr2026_conference,times}

\usepackage{amsmath,amsfonts,bm}

\def\eqref#1{equation~\ref{#1}}
\def\1{\bm{1}}

\DeclareMathAlphabet{\mathsfit}{\encodingdefault}{\sfdefault}{m}{sl}
\SetMathAlphabet{\mathsfit}{bold}{\encodingdefault}{\sfdefault}{bx}{n}

\usepackage{hyperref}
\usepackage{url}
\usepackage{array}
\usepackage{tabularx}
\usepackage{multirow}
\usepackage{booktabs}
\usepackage{hyperref}
\usepackage{pifont}
\usepackage{graphicx}
\usepackage{subcaption}
\usepackage[table]{xcolor}
\usepackage{wrapfig}
\usepackage{siunitx}
\usepackage{array}
\usepackage{booktabs}
\usepackage{authblk}
\usepackage{amssymb}

\definecolor{ModelGreen}{RGB}{213,232,212}

\title{~\includegraphics[height=25pt]{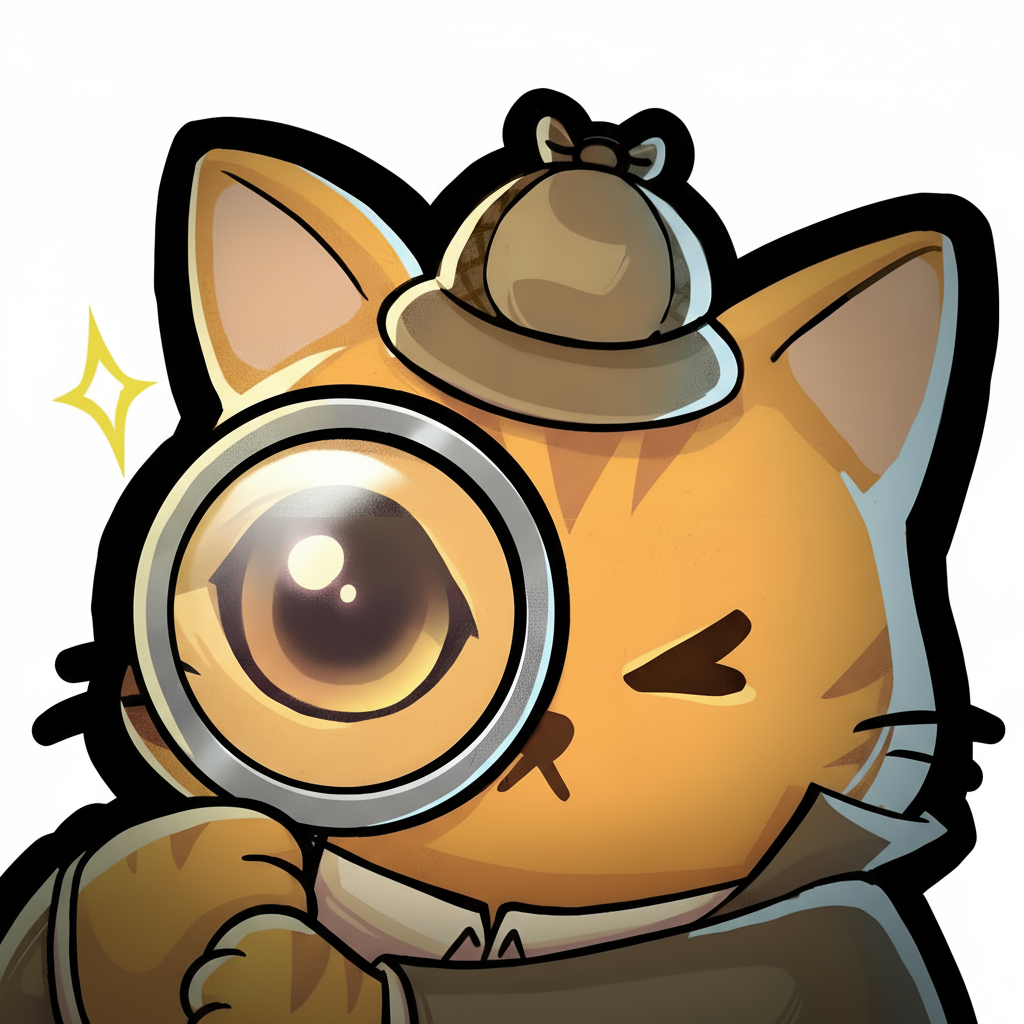}Video Detective: Seek Critical Clues Recurrently to Answer Question \\ from Long Videos
}

\author{
\textbf{Henghui Du}\textsuperscript{1,2,\dag},
\textbf{Chunjie Zhang}\textsuperscript{2},
\textbf{Xi Chen}\textsuperscript{2},
\textbf{Chang Zhou}\textsuperscript{2,*},
\textbf{Di Hu}\textsuperscript{1,*}
\\
\textsuperscript{1}Gaoling School of Artificial Intelligence, Renmin University of China, Beijing\\
\textsuperscript{2} AI Technology Center, Online Video Business Unit, Tencent PCG \\
\tt\small{\textsuperscript{1}\{cserdu,dihu\}@ruc.edu.cn} \\
\tt\small{\textsuperscript{2}\{cserdu,chanzhou,esmezhang,jasonxchen\}@tencent.com}
}

\iclrfinalcopy % Uncomment for camera-ready version, but NOT for submission.
\begin{document}

\maketitle
\footnotetext{\textsuperscript{\dag} Intern at Tencent PCG}
\footnotetext{\textsuperscript{*} Co-corresponding author}

\begin{abstract}
Long Video Question-Answering (LVQA) presents a significant challenge for Multi-modal Large Language Models (MLLMs) due to immense context and overloaded information, which could also lead to prohibitive memory consumption. 
While existing methods attempt to address these issues by reducing visual tokens or extending model's context length, they may miss useful information or take considerable computation.
In fact, when answering given questions, only a small amount of crucial information is required.
Therefore, we propose an efficient question-aware memory mechanism, enabling MLLMs to recurrently seek these critical clues. Our approach, named VideoDetective, simplifies this task by iteratively processing video sub-segments. For each sub-segment, a question-aware compression strategy is employed by introducing a few special memory tokens to achieve purposefully compression. This allows models to effectively seek critical clues while reducing visual tokens.
Then, due to history context could have a significant impact, we recurrently aggregate and store these memory tokens to update history context, which would be reused for subsequent sub-segments. 
Furthermore, to more effectively measure model's long video understanding ability, we introduce GLVC (Grounding Long Video Clues), a long video question-answering dataset, which features grounding critical and concrete clues scattered throughout entire videos.
Experimental results demonstrate our method enables MLLMs with limited context length of $32K$ to efficiently process $100K$ tokens ($3600$ frames, an hour-long video sampled at $1fps$), requiring only $2$ minutes and $37$GB GPU memory usage. Evaluation results across multiple long video benchmarks illustrate our method can more effectively seek critical clues from massive information.
Code and dataset: \href{https://github.com/CserDu/VideoDetective}{https://github.com/CserDu/VideoDetective}
\end{abstract}

\section{Introduction}
\label{sec:intro}

% As an indispensable part of human daily life, movies are ubiquitous in our lives. As Vladimir Lenin said, \textit{``That of all the arts, the most important for us is the cinema."} Movies are life and it is of great significance to teach machines to understand movies. In recent years, Multi-modal Large Language Models (MLLMs) for video understanding have made significant progress~\citep{cheng2024videollama,li2024llama,bai2025qwen2,li2023videochat}. However, unlike ordinary videos, movies are usually minutes-long or even hours-long videos, and they have complex character relationships and high-level plots compared with basic visual concept. Some recent research works~\citep{han2024autoad,ji2024ida,he2024storyteller,weng2024longvlm,song2024moviechat} have focused on understanding movie videos. They either solve long video understanding tasks with pure visual content~\citep{weng2024longvlm,song2024moviechat}, or consider characters in seconds-long videos~\citep{han2024autoad,ji2024ida}. There is still a lot of exploration space for the understanding of movie videos.

Long video question-answering task aims to answer questions from minutes-long or even hours-long videos, requiring models to process a large number of video frames. Mainstream Multi-modal Large Language Models (MLLMs) represent a single video frame with a substantial number of tokens. For example, Qwen2.5-VL~\citep{bai2025qwen2} represents an image of $224\times224$ resolution as $64$ tokens, while LLaVA-NeXT~\citep{zhang2024llavanext-video} uses $144$ tokens. 
% Therefore, a long video input could result in extremely long token sequences, thus easily exceeding the limits of MLLMs’ context lengths and causing memory explosion.
Therefore, a ten minutes-long video input ($600$ frames sampled at $1fps$) could result in at least $38K$ tokens, which easily exceeds the maximum context length of Qwen2.5-VL ($32K$) and causes memory explosion.

% take considerable computation and GPU memory costs.

% the input may easily exceed the limits of MLLMs’
% context lengths

% For instance, qwen2.5-vl represents a 224x224 resolution image as 64 tokens, while llava-next uses 144 tokens, and so on.

% Long video question-answering task aims to answer questions from minutes-long or even hours-long videos, which is extremely difficult for existing Multi-modal Large Language Models (MLLMs) due to extremely long context and massive information. 
% Mainstream MLLMs represent a video frame with a large number of visual tokens (i. e., 64 tokens for Qwen2.5-VL and 144 tokens for LlaVA-NeXT, ect.), which can only process 600 frames with 64GB GPU memory.
% A long video consists of a long sequence of frames, each of which usually consumes a large number of visual tokens for MLLMs to perceive (e.g., 144 tokens per frame). 
% For example, for the Qwen2.5-VL model~\citep{bai2025qwen2} which represents each frame with 64 tokens, it can only process 600 frames with 64GB GPU memory, as shown in Fig.~\ref{fig:teaser}(b).
% Recently, there have been many research works dedicated to solving this problem. They either extended the context length of MLLMs~\citep{dubey2024llama,bai2023qwen,xiong2023effective} but at the cost of considerable computational resources, or reducing the overall token numbers by merging adjacent frames with similar semantics~\citep{he2024ma,song2024moviechat,weng2024longvlm}, which could lead to a substantial loss of visual information.

Recently, there have been many research works dedicated to solving these problems.
% by reducing the number of visual tokens~\citep{he2024ma,song2024moviechat, weng2024longvlm} or extending the model's context length.
% The former
They either reduced the number of visual tokens by merging adjacent frames with similar semantics~\citep{he2024ma,song2024moviechat,weng2024longvlm} or extended the context length of MLLMs followed by fine-tuning on the long videos data ~\citep{dubey2024llama,bai2023qwen,xiong2023effective}. While these approaches enable handing longer video input, the former may result in the loss of significant information, affecting the understanding of long videos; the latter takes considerable computation and GPU memory costs. 
% In this work, we aim to propose a more efficient method.
Therefore, a more efficient method is urgently needed for long video understanding task.

% \begin{figure*}[!t]
%      \centering
%      \includegraphics[width=0.98\textwidth]{iclr2026/figs/iclr-figs/teaser-a.png}
%      \vspace{-1em}
%      \caption{We present VideoDetective, an efficient Multi-modal Large Language Model, capable of answering question from minutes-long or even hours-long videos by \textbf{recurrently seeking critical clues related to question.} As shown in (a), it processes multiple video sub-segments recurrently. In each iteration, few special tokens are introduced to seek semantic representations related to question from history context and current inputs.Then these tokens are aggregated and stored to update history context. Finally, compared to entire video, only extremely few tokens are used to answer question, thus saving memory usage while effectively seeking critical clues. Fig.(b) shows the evaluation results compared with other models. Our method achieves \textbf{superior results with comparable inference time but much lower GPU memory usage.}
%     }
%      \label{fig:teaser}
%      \vspace{-1.25em}
% \end{figure*}

% \begin{figure*}[!t]
%      \centering
%      \includegraphics[width=0.98\textwidth]{iclr2026/figs/iclr-figs/teaser-b.png}
%      \vspace{-1em}
%      \caption{Fig (a). radar. Fig (b) inference efficiency}
%      \label{fig:teaser}
%      \vspace{-1.25em}
% \end{figure*}

\begin{figure*}[!t]
    \centering
    
    \begin{subfigure}{\linewidth}
        \centering
        \includegraphics[width=0.95\linewidth]{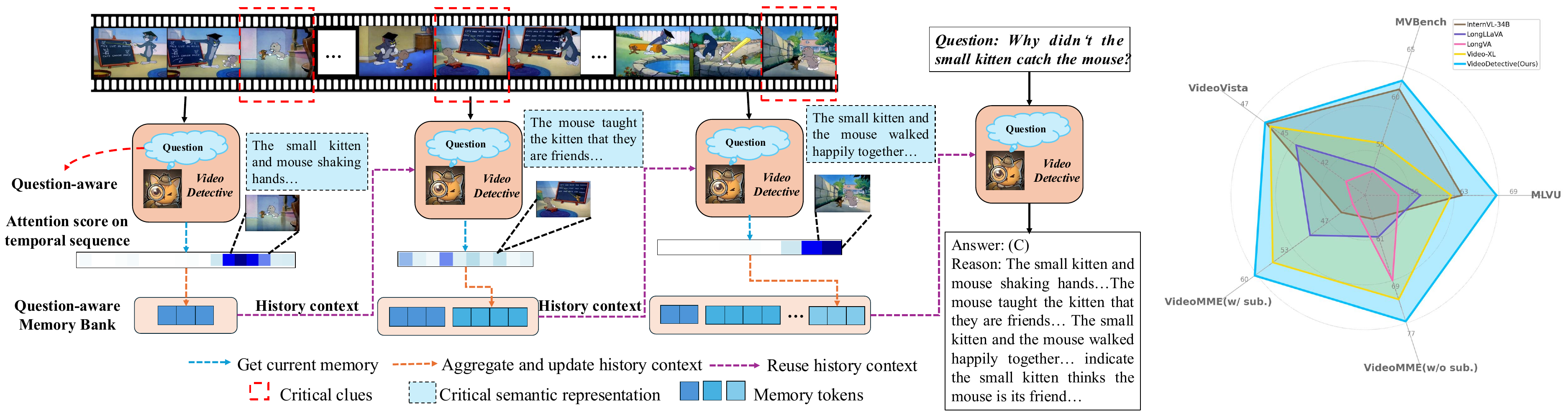}
        \caption{~Left: The overview of VideoDetective equipped with an efficient question-aware memory mechanism. Right: Evaluation results on multiple long video benchmarks.}
        \label{fig:teaser-a}
    \end{subfigure}
    
    % \vspace{1em} % 添加垂直间距
    
    % \begin{subfigure}{0.43\linewidth}
    %     \centering
    %     \includegraphics[width=\linewidth]{iclr2026/figs/iclr-figs/radar.png}
    %     \caption{fig b caption}
    %     \label{fig:teaser-b}
    % \end{subfigure}
    % \hfill
    % \begin{subfigure}{0.43\linewidth}
    %     \centering
    %     \includegraphics[width=\linewidth]{iclr2026/figs/iclr-figs/teaser-c.png}
    %     \caption{fig c caption}
    %     \label{fig:teaser-c}
    % \end{subfigure}

    \begin{subfigure}{\linewidth}
        \centering
        \includegraphics[width=0.95\linewidth]{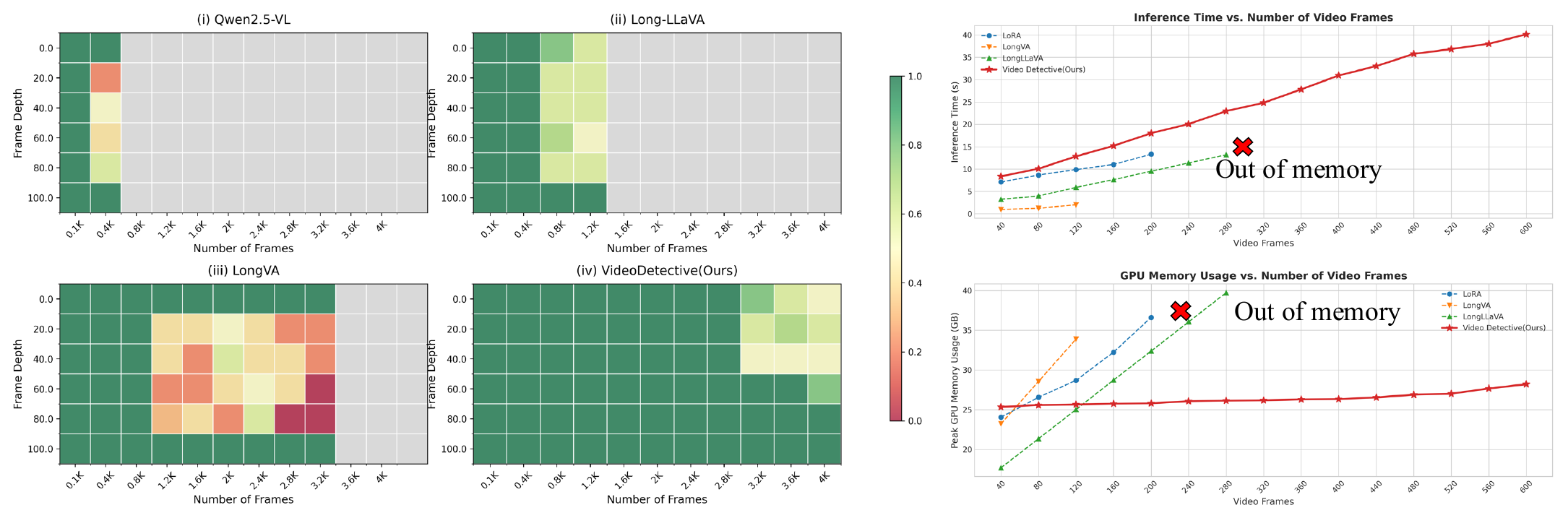}
        \caption{~Left: Visual Needle-In-The-Haystack evaluation. The x-axis represents the total number of frames in the video haystack. The y-axis shows the position where the needle image is located. Gray grids mean “OOM’. Right: Inference efficiency compared with LoRA and two long video models.}
        \label{fig:teaser-bc}
    \end{subfigure}
    
    \caption{We present VideoDetective, an Multi-modal Large Language Model equipped with efficient question-aware memory mechanism. As shown in Fig.~\ref{fig:teaser-a}, it features \textbf{recurrently seeking critical clues related to question} from minutes or even hours-long videos. Compared to entire videos, only a few special tokens are used to answer question, thus saving GPU memory usage while effectively leveraging crucial information.
    % As shown in Fig.(a), it processes multiple video sub-segments recurrently. In each iteration, few special tokens are introduced to seek semantic representations related to question from history context and current inputs.Then these tokens are aggregated and stored to update history context. Finally, compared to entire video, only extremely few tokens are used to answer question, thus saving memory usage while effectively seeking critical clues.
    Fig.~\ref{fig:teaser-bc} shows the Visual Needle-In-The-Haystack evaluation~\citep{zhang2024long} and inference efficiency. Our proposed efficient question-aware memory mechanism enables models with limited context length, such as Qwen2.5-VL, to efficiently process $4K$ video frames input, requiring only $2$ minutes and $37$GB GPU memory usage. Moreover, compared to other long video understanding models, VideoDetective could more effectively seek critical ``needles" from video haystack,
    demonstrating superior long video understanding capabilities.
    }
    \label{fig:teaser}
    \vspace{-2em}
\end{figure*}

% Surprisingly, it can \textbf{process an hour-long video at 1fps (3600 frames) while effectively seeking critical clues, which only requires 2 minutes and 37GB memory usage.

% The Shawshank Redemption
% As a type of video frequently encountered in people's daily lives, movies are artificially edited and more complex long videos. 

% Essentially, most of information is redundant and irrelevant to the question for a long video. 
Essentially, a long video contains a massive amount of information, most of which is redundant and irrelevant to the question, 
% thereby increasing the difficulty for the model to comprehend and complete  task.
% Conversely, our approach focuses on identifying a small amount of key information, offering a more efficient method.
Thus, it is a more efficient approach is to seek small amount of crucial clues.
% When answering given questions, actually, only a small amount of crucial information is required. 
For example, as shown in Fig.~\ref{fig:teaser-a}, in the cartoon video of ``Tom and Jerry", there are three critical clues indicating that the small kitten and Jerry are friends (as marked by the red dashed line). Only combining these clues, the model can correctly answer the question ``Why didn't the small kitten catch the mouse?"
% Through experimental observation we found that it is difficult for model to effectively find crucial information.
% As shown in Fig.~\ref{fig:lora-attn}, it displays the attention scores of last layer in LLMs.
% Compared with 5 frames, the model's attention to entire video (120 frames) is usually indiscriminate, and there are no prominent visual tokens it focuses on. 
% In this case, to correctly answer question, it is necessary to find important information from the long context, which is akin to searching for a needle in a haystack.
More intuitively, when we humans complete this task, we usually adopt a strategy of \textbf{thinking while watching}, that is, thinking with questions in mind. During this process, we look for important information related to question and remember them in our mind. Then we continue to watch subsequent video content until end. Finally we integrate all collected information in the past to answer question. Compared with watching entire video at once and then thinking about question, this strategy is easier and more efficient.

Inspired by this progressive thinking process, we propose an efficient question-aware memory mechanism, enabling models to recurrently seek critical clues related to question from long videos.
% Specifically, we first simplify this complex task by dividing a long video into multiple short sub-segments and processing them recurrently. In each iteration, we use few special tokens to seek semantic representation related to question from history context and current sub-segment. Then these tokens are aggregated and stored to update history context. Finally, only extremely few tokens are used to answer question, thus saving memory usage while effectively seeking critical clues. 
% Specifically, due to the causal attention and autoregressive na-
% ture, the language models will spontaneously aggregates the sequence information onto the last few
% tokens (Lester et al., 2021; Kitaev et al., 2020). These tokens naturally serve as a compact represen-
% tation and provide a high-level summary of current input sequence.
Specifically, it consists of two parts: \textbf{1) question-aware compression of visual tokens}:
% leverage the causal attention and autoregressive properties of language models~\citep{lester2021power,kitaev2020reformer}
we first append a few learnable special tokens (also called memory tokens) to the end of each input video sub-segment. To avoid missing crucial information, the questions are inserted before memory tokens to achieve purposeful compression. Therefore, these memory tokens naturally serve as queries and could aggregate crucial semantic representations due to the causal attention and autoregressive properties of language models~\citep{lester2021power,kitaev2020reformer}.
\textbf{2) recurrently seek critical clues}: as history context could also have a significant impact on critical clues, the memory tokens with crucial semantic from each sub-segment are extracted and stored in the memory bank to update history context. They would be reused as additional inputs when processing subsequent sub-segments.
Finally, only extremely few memory tokens are used to answer the question, without relying on any video input, thus saving GPU memory usage while effectively seeking critical clues.
% to ensure the integrity of global context, we iteratively process these video sub-segments. In each iteration, the semantic representations are aggregated and then stored in a memory bank to update the historical semantics.
% Evaluating the model’s ability to understand long videos is of significant importance. Although there
% have been many long video benchmarks available, they mainly assess the model’s final prediction
% results, which can not guarantee the model truly understands long videos. Inspired by the Visual
% Needle-In-The-Haystack evaluation (Zhang et al., 2024a), we further propose the GLVC (Grounding
% Long Video Clues), a long video question-answering dataset that features grounding critical and
% concrete clues scattered throughout entire videos.

Further, to more effectively measure the model’s ability to seek critical clues from long videos, we introduce \textbf{GLVC} (\textbf{G}rounding \textbf{L}ong \textbf{V}ideo \textbf{C}lues), a long video question-answering dataset that features grounding concrete and critical clues scattered throughout entire long videos.
Although there have been many long video benchmarks available, the tasks in them can be completed relying solely on a small set of sampled video frames, and they mainly evaluate model's final prediction results.
Different from them, GLVC includes concrete crucial clues and timestamps (as shown in Fig.~\ref{fig:dataset-example}), which can be used to quantitatively assess whether the models truly understand long videos.
Experimental results across multiple benchmarks, as shown in Fig.~\ref{fig:teaser-a} (Right) demonstrate our method achieves superior results. Further, the evaluation results on GLVC dataset illustrate it has stronger capability to seek critical clues compared to other long video understanding models.Overall, our contributions can be summarized as follows.

% the task can be completed with a small number of sampled video frames, and evaluation is performed only on the model's final prediction results.

% Overall, our contributions can be summarized as follows.
\begin{itemize}
    \item We propose an efficient question-aware memory mechanism to recurrently seek critical clues related to question from long videos, enabling models with limited context length of $32K$ to efficiently process $100K$ tokens (an hour-long video sampled at $1fps$), only requiring $2$ minutes and $37$GB GPU memory usage.
    \item To more effectively measure model's ability to seek critical clues from long context, we introduce GLVC (Grounding Long Video Clues), a long video question-answering dataset. Different from existing long video benchmarks, it features grounding critical and concrete clues scattered throughout entire videos.
    \item Experimental results across multiple long video benchmarks show that our method can significantly reduce GPU memory usage with comparable inference time while effectively seeking critical clues from long videos, demonstrating great potential on hours-long video understanding tasks. 
\end{itemize}

% \emph{Looking for answers with questions while watching movies is much easier than thinking about the answer to the question after watching the entire movie.} 

% \begin{figure*}[!t]
%      \centering
%      \includegraphics[width=0.95\textwidth]{iclr2026/figs/dataset.pdf}
%      \vspace{-1em}
%      \caption{
%      The overview of our M3QA dataset. (a) The dataset construction pipeline. (b) The video segment duration statistics. (c) The data structure division in M3QA.
%      }
%      \label{fig:dataset}
%      \vspace{-1.25em}
% \end{figure*}

\section{Related Work}
\label{sec:related_work}

\subsection{\textbf{MLLMs for Short Video Understanding}}
In the past few years, MLLMs for short video understanding have made significant progress~\citep{cheng2024videollama,li2023videochat,ataallah2024minigpt4,sun2024video,lin2023video}. VideoLLaMA 2~\citep{cheng2024videollama} incorporates a tailor-made spatial-temporal convolution connector to effectively capture the intricate spatial and temporal dynamics of video data. Video-LLaVA~\citep{lin2023video} aligns multi-modal representations before projection and endows LLM with the ability to comprehend both images and videos simultaneously.
To obtain fine-grained temporal information required by video understanding, Sun \emph{et al.}~\citep{sun2024video} propose a multi-resolution causal Q-Former structure. 
% Qwen2.5-VL~\citep{bai2025qwen2} integrates multimodal rotary position embedding(M-RoPE) to facilitate the effective fusion of positional information across text, image and videos. 
These works mainly focus on short video understanding, and the models have difficulty in handling long videos.

\subsection{\textbf{Long Video Question Answering}}
Originating from long text modeling in natural language processing, some research works~\citep{song2024moviechat,weng2024longvlm,li2024llama,qian2025streaming} have been attempted to solve the long video question answering task. Song \emph{et al.}~\citep{song2024moviechat} and Weng \emph{et al.}~\citep{weng2024longvlm} merge the most similar tokens in the adjacent frames to reduce token numbers and enable LLMs to process long videos. LLaMA-VID~\citep{li2024llama} represents each frame with two distinct tokens, namely context token and content token to reduce the computational overhead while preserving the critical information. 
Qian \emph{et al.}~\citep{qian2025streaming} sequentially encodes video clips and distills condensed representation using a constant number of video tokens.
While these works can effectively reduce token numbers, they usually adopt a purposeless compression method, which may lose important information. Unlike them, our method can effectively seek critical information related to question using only few tokens.

% \subsection{\textbf{MLLMs for Movie Understanding}}
% As a type of long video frequently encountered in people's daily life, movies have rich characters and substantial plots. Enabling machines to understand movies like humans is a challenging task, which has gradually attracted the interest of researchers. Han \emph{et al.}~\citep{han2023autoad,han2024autoad} and Ji \emph{et al.}~\citep{ji2024ida} incorporate character information based on visual concept, allowing the model to describe plots related to characters.
% Huh \emph{et al.}~\citep{huh2024character} integrates speech recognition, speaker diarisation to generate subtitles with speaker. StoryTeller~\citep{he2024storyteller} considers characters and speech content in fixed 3 minutes-long videos to complete plot understanding. 
% While they have made some progress, there is still a lack of sufficient exploration in answering questions from minute-long or even hours-long movie videos.

\subsection{\textbf{Long Video Understanding Benchmarks}}
% Traditionally, MLLMs's video understanding ability is evaluated on video QA datasets like ActivityNet-QA
% For long video understanding, researchers leveraged long-form videos, like movies, to create benchmarks. For example, the LVU dataset~\citep{zhou2024mlvu} presents multiple movie understanding tasks, such as predicting release years and ide
For long video understanding, researchers have proposed numerous benchmarks~\citep{wang2024lvbench, zhou2024mlvu,wu2024longvideobench, fu2025video} from various perspectives to evaluate the model's ability. For example, the LVU dataset~\citep{zhou2024mlvu} collected movie videos with varying lengths and constructed diversified tasks to comprehensively evaluation the model's ability.
LongVideoBench~\citep{wu2024longvideobench}, includes varying-length web-collected videos with subtitles across diverse themes, featuring video-language interleaved inputs up to an hour for longer and richer inputs. However, these benchmarks simply evaluate model's final prediction results, which can not comprehensively prove the model truly understands long videos. In this work, we further propose the GLVC dataset, which could be used to more effectively measure model's ability to seek critical clues from long context.

\section{Method}
\label{sec:method}

\begin{figure*}[t]
     \centering
     \includegraphics[width=0.97\textwidth]{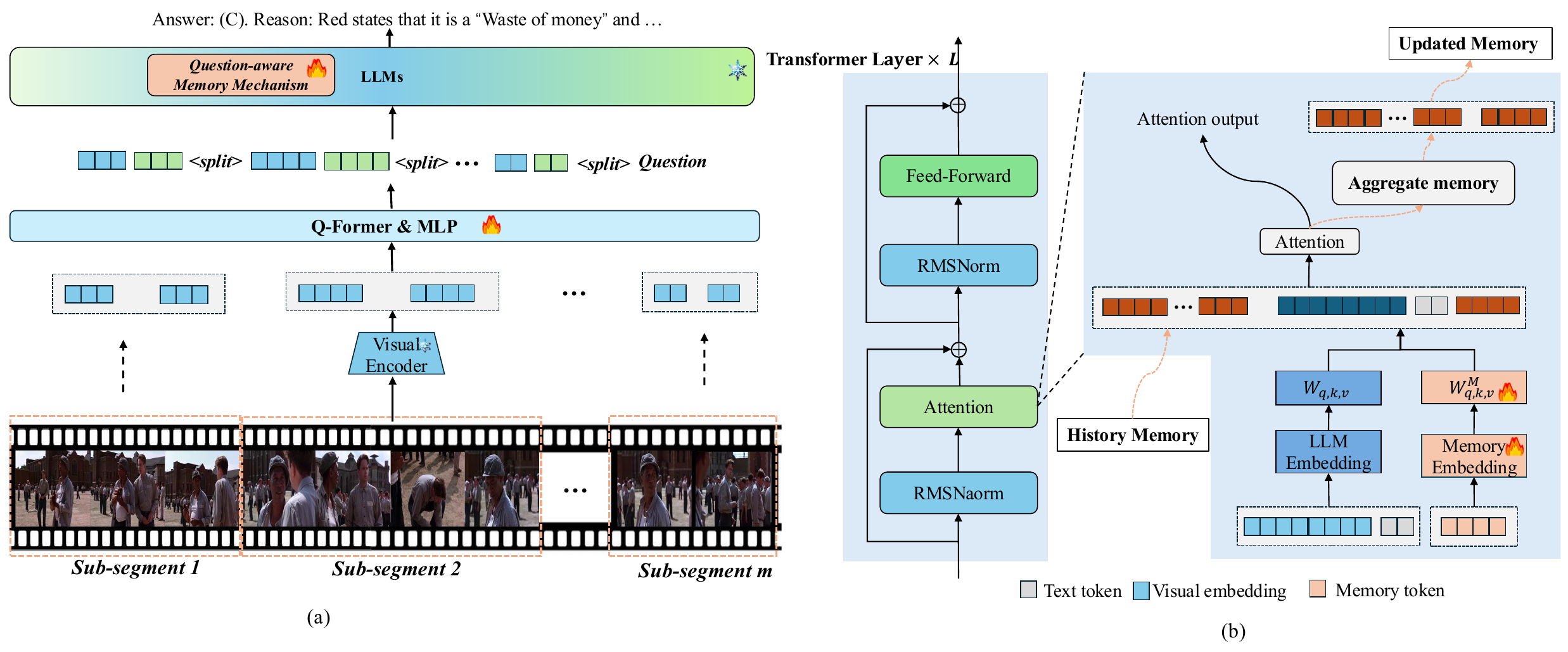}
     \vspace{-1em}
     \caption{
     (a) The architecture of our VideoDetective model. The video segment is divided into multiple sub-segments, which are processed by visual encoder to get multi-modal embeddings. Then these embeddings are separated by special \texttt{<split>} tokens and the model processes each sub-segment recurrently. (b) The question-aware memory mechanism in attention module at every transformer layer of LLMs. During the process of processing each sub-segment, only few special memory tokens \texttt{<memory>} are appended at the end of sub-segment sequence. Then the memory tokens from all past sub-segments and current sub-segment perform attention calculations with other tokens, and then all the memory tokens are aggregated and stored as historical information for subsequent sub-segments.
     }
     \label{fig:framework}
     \vspace{-1.25em}
\end{figure*}

% \begin{figure*}[t]
%     \vspace{-1em}
%   \centering
%   \begin{subfigure}[b]{0.47\textwidth}
%     \includegraphics[width=\textwidth]{figs/framework-a.png}
%     \caption{The overview of framework.}
%     \label{fig:framework-a}
%   \end{subfigure}
%   \hfill % Optional: adjust the horizontal spacing between the subfigures
%   \begin{subfigure}[b]{0.47\textwidth}
%     \includegraphics[width=\textwidth]{figs/framework-b.png}
%     \caption{The question-aware memory mechanism at every transformer layer.}
%     \label{fig:framework-b}
%   \end{subfigure}
%   \vspace{-0.5em}
%   \caption{Our framework.}
%   \label{fig:framework}
%   \vspace{-1.5em}
% \end{figure*}

% In this section, we first introduce the task overview in Sec.~\ref{method:task-overview}, and then describe our model architecture in Sec.~\ref{method:multimodal-branch}. Subsequently, in Sec.~\ref{method:memory} and Sec.~\ref{method:train+infer} we provide a detailed introduction about question-aware memory mechanism, its training and inference process.

In this section, we first introduce the model architecture of VideoDetective in Sec.~\ref{method:arch}, then we provide a detailed introduction about question-aware memory mechanism in Sec.~\ref{method:memory}, followed by the training and inference process in Sec.~\ref{method:train+infer}
.
% \subsection{Task Overview}
% \label{method:task-overview}

% \begin{equation}
%     P(\mathcal{O}|\mathcal{V},\mathcal{A},\mathcal{C},\mathcal{Q}) = P(\mathcal{O}|\mathcal{V}_{sub},\mathcal{A}_{sub},\mathcal{C},\mathcal{Q})
% \end{equation}

\subsection{Model Architecture}
\label{method:arch}
The model architecture of VideoDetective follows a standard design, which includes a visual encoder, a visual-language projector, and an LLM backbone equipped with an efficient question-aware memory mechanism, as shown in Fig.~\ref{fig:framework}(a). 
Given a question $\mathcal{Q}$ and a long video input $\mathcal{I}=\{I_{i} \in \mathbb{R}^{H \times W \times C}\}_{i=1}^{T}$, where $H, W, C$ represent the height, width and channels respectively, $T$ is the number of video frames, the model needs to predict the answer, denotes as $\mathcal{A}$. The long video is first divided into $S$ sub-segments $\mathcal{V} = \{ {V}_i \}_{i=1}^{S}$.
% Then these embeddings are separated by special \texttt{<split>} tokens and the model processes each sub-segment recurrently.
Then we expand the tokenizer by adding an additional special token \texttt{<split>}, which is inserted between adajcent video sub-segments to facilitate distinction.
The efficient question-aware memory mechanism is applied in the attention modules of each transformer layer of the LLM backbone, as shown in Fig.~\ref{fig:framework}(b).
Next, we will provide a detailed introduction to this module.

\subsection{Efficient question-aware memory mechanism}
\label{method:memory}
\textbf{Question-aware Compression of Visual Tokens} 
% The answer to question is usually related to few important and scattered information. Huge amount of irrelevant plots make it difficult for model to find these information. To simplify this complex task, we first divide the video segment into multiple sub-segments. 
% Fixed-length sub-segments could disrupt the complete semantics of original shots in the movie. Therefore, we use the sub-segments already divided in M3QA dataset.
% Intuitively, when we humans think about question, we often watch video with question, and obtain critical clues related to the question during watching, which is easier than watching entire video and then thinking about question. 
% Intuitively, \emph{watching a video with question and collecting important information during watching is simpler than watching the entire video and then thinking about the question.}
% In order to reduce the number of visual tokens, it is essential to compress token numbers in each sub-segment.
Due to the causal attention and autoregressive nature, the language models will spontaneously aggregates the sequence information onto the last few tokens~\citep{lester2021power,kitaev2020reformer}. These tokens naturally serve as a compact representation and provide a high-level summary of current input sequence. 
Therefore, we can introduce a small set of learnable memory tokens $M = \{\texttt{<memory>}_1, \cdots, \texttt{<memory>}_k\}$, where $k$ is the number of memory tokens. Then these tokens are appended to the end of the video segments to compress visual tokens.
To avoid the loss of important information and achieve purposeful compression, we additionally insert the questions into current inputs to facilitate seeking critical clues.
Therefore, the inputs are organized as:
\begin{gather}
    \underbrace{ \{V_1, \mathcal{Q},  M_1\}}_{Seg_1}, \texttt{<split>} , \underbrace{ \{V_2, \mathcal{Q},  M_2, \}}_{Seg_2}, \texttt{<split>}, \cdots, \underbrace{ \{V_s, \mathcal{Q},  M_s\}}_{Seg_S}, \texttt{<split>}, \mathcal{Q},
\end{gather}
where $M_i$ is the memory tokens for video segment $V_i$.
Considering the video sub-segments of different lengths may contain varying amounts of information, thus requiring storage of more critical clues, we introduce a compression ratio $\alpha$ to dynamically adjust the number of memory tokens.
It can be formalized as $k = N_i / \alpha$, where $N_i$ is the visual token numbers of $V_i$.

% However, the model still needs to remember all the information of current sub-segment through these tokens, which is a purposeless information compression approach.
% Intuitively, \emph{watching video with question and seeking important information during watching is simpler than watching entire video and then thinking about question.} When we humans think about questions, we often watch videos with questions, and mainly focus on critical information related to question during watching. 

\begin{wrapfigure}[10]{r}{0.46\textwidth}
    \centering
    \includegraphics[width=0.40\textwidth]{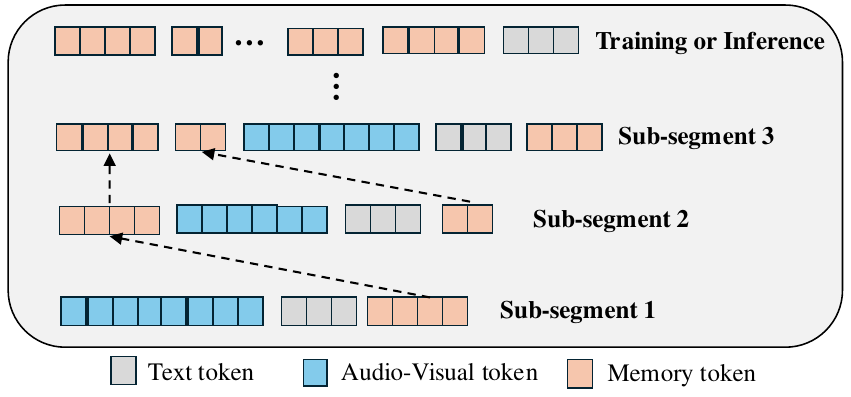}
    \caption{The training and inference process.}
    \label{fig:train+infer}
\end{wrapfigure}

\textbf{Recurrently Seek Critical Clues}
% Inspired by this progressive thinking process, we propose an efficient question-aware memory mechanism to recurrently seek critical clues related to question from long videos. 
% Specifically, we first simplify this complex task by dividing video into multiple sub-segments and processing them recurrently. During this process,  few special memory tokens $\texttt{<memory>}$ are introduced and appended at the end of current sub-segment. To effectively seek semantic representation related to question, we further allow these tokens pay attention to the question. Considering impact of history context on the semantics, all past memory tokens are also aggregated as current inputs. 
% Therefore, when processing each sub-segment, inputs are organized as:
% \begin{gather}
%     \{V_1, \mathcal{Q},  M_1\}, \texttt{<split>},\{V_2, \mathcal{Q},  M_2, \}, ..., \{V_s, \mathcal{Q},  M_s\}, \texttt{<split>}, \mathcal{Q}
% \end{gather}
A long video is split into $S$ sub-segments, which are processed recurrently. In each iteration, the model needs to seek and preserve critical semantic representations related to question, which are then input as history context for subsequent processing steps. Specifically, as shown in Fig.~\ref{fig:framework}(b), assume in the $t_{th}$ iteration, in each transformer layer of LLM backbone, the memory tokens $M_t$ for current video segments $V_t$ are first converted into memory embedding, and then multiplied with three trainable matrices $W_{q,k,v}^{m}$ to obtain $Q^m$, $K^m$, and $V^m$:
\begin{gather}
    Q^m, K^m, V^m = W_{q,k,v}^m \cdot E^m(M_t),
\end{gather}
where $E^m(\cdot)$ is a trainable token embedding similar to word embedding in LLMs.
The $Q$, $K$, $V$ for video and question embeddings can also be obtained in the same way:
\begin{gather}
    Q, K, V = W_{q,k,v} \cdot g([F_v^t \circ F_{\mathcal{Q}}]),
\end{gather}
where $g(\circ)$ denotes concatenation operation.
The $query$, $key$, and $value$ in self-attention calculation can be denoted as:
\begin{gather}
    query = g([Q \circ Q^m]),\\
    key = g([F_k^{past} \circ K \circ K^m]),\\
    value = g([F_v^{past} \circ V \circ V^m]),
\end{gather}
where $F_k^{past}, F_v^{past}$ are history context stored in the memory bank from previous iterations.
Finally, we aggregate and extract the semantic representations corresponding to memory tokens from results of self-attention computation, which include critical clues related to question. These representations are stored in memory bank to update history context. 

Since above process occurs in each transformer layer, the memory bank stores history context from all layers, which are then extracted and reused to each layer in subsequent iterations. In the last step of iteration, the model predicts the answer only based on the history context stored in the memory bank, without relying on any video inputs.

% \begin{figure}[t]
%      \centering
%      \includegraphics[width=0.47\textwidth]{iclr2026/figs/train+infer.pdf}
%      \vspace{-1em}
%      \caption{
%      The training and inference process.
%      }
%      \label{fig:train+infer}
%      \vspace{-1.25em}
% \end{figure}

\subsection{Training and Inference Process}
\label{method:train+infer}
Fig.~\ref{fig:train+infer} demonstrates the overview of training and inference process. 
As mentioned in sec~\ref{method:memory}, the number of memory tokens will gradually accumulate with the iterations progress.
% Since the number of memory tokens in each sub-segment is smaller than audio-visual tokens, the tokens in each sub-segment will not increase much, and the total number of memory tokens is far less than the tokens of the entire video segment. 
% \begin{gather}
%     \mathcal{L} = \sum_{i=1}^{S} \sum_{j=1}^{l} -log\ p(x_j | M_1,...,M_{i-1},x_0,...,x_{j-1}),
% \end{gather}
% where $S$ is the sub-segment numbers, $l$ is the token numbers of instruction.
During training process, memory tokens do not participate in loss calculation, and they are only used as an implicit semantic representation related to question. Therefore, the training loss only consists of the cross-entropy loss:
\begin{gather}
    \mathcal{L} = \sum_{j=1}^{l} -log\ p(x_j |V_1, M_1, \mathcal{Q}, \cdots ,V_S, M_{S}, \mathcal{Q},x_0,...,x_{j-1}),
\end{gather}
where $l$ is the token numbers of output $\mathcal{O}$.

During the inference process, all sub-segments are processed in the same way as the training process. After all sub-segments are processed, the question and all memory tokens are used as input to predict the answer, without any video inputs. The next token is predicted as:
\begin{gather}
     p(x_j | M_1,...,M_{S},x_0,...,x_{j-1}),
\end{gather}

% \begin{figure*}[ht]
%      \centering
%      \includegraphics[width=0.95\textwidth]{figs/framework.png}
%      \vspace{-1em}
%      \caption{
%      Our model framework.
%      }
%      \label{fig:framework}
%      \vspace{-1.25em}
% \end{figure*}

% \begin{figure}[ht]
%      \centering
%      \includegraphics[width=0.5\textwidth]{figs/memory.png}
%      \vspace{-1em}
%      \caption{
%      Memory structure.
%      }
%      \label{fig:memory}
%      \vspace{-1.25em}
% \end{figure}

\section{GLVC Dataset}
\label{sec:dataset}
Evaluating the model's ability to understand long videos is of significant importance. Although there have been many long video benchmarks available, they mainly assess the model's final prediction results, which can not guarantee the model truly understands long videos. Inspired by the Visual Needle-In-The-Haystack evaluation~\citep{zhang2024long}, we further propose the GLVC (\textbf{G}rounding \textbf{L}ong \textbf{V}ideo \textbf{C}lues), a long video question-answering dataset that features grounding critical and concrete clues scattered throughout entire videos. In this section, we first introduce the construction pipeline of GLVC dataset in Sec.~\ref{sec:dataset-construct} and then provide a detailed analysis in Sec.~\ref{dataset:analysis}.

% \begin{figure*}[!t]
%      \centering
%      \includegraphics[width=0.95\textwidth]{iclr2026/figs/iclr-figs/dataset.png}
%      \vspace{-1em}
%      \caption{
%      The dataset construction pipeline of GLVC dataset.
%      }
%      \label{fig:dataset}
%      \vspace{-1.25em}
% \end{figure*}

\begin{figure*}[!t]
    \centering
    \begin{subfigure}{\linewidth}
        \centering
        \includegraphics[width=0.95\linewidth]{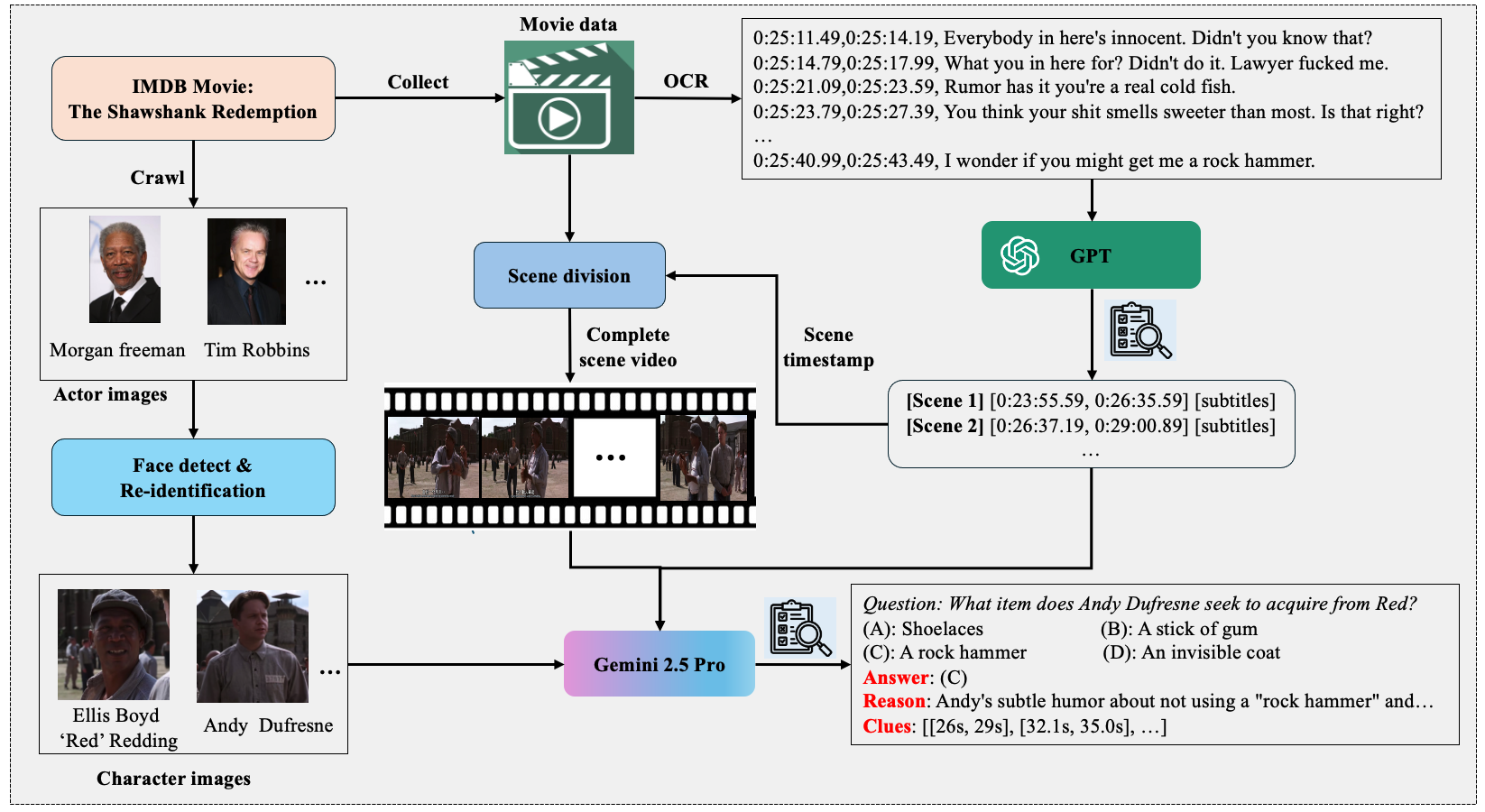}
        \caption{~The detailed construction pipeline of GLVC dataset.}
        \label{fig:dataset-construct}
    \end{subfigure}
    
    % \vspace{1em} 

    \begin{subfigure}{\linewidth}
        \centering
        \includegraphics[width=0.95\linewidth]{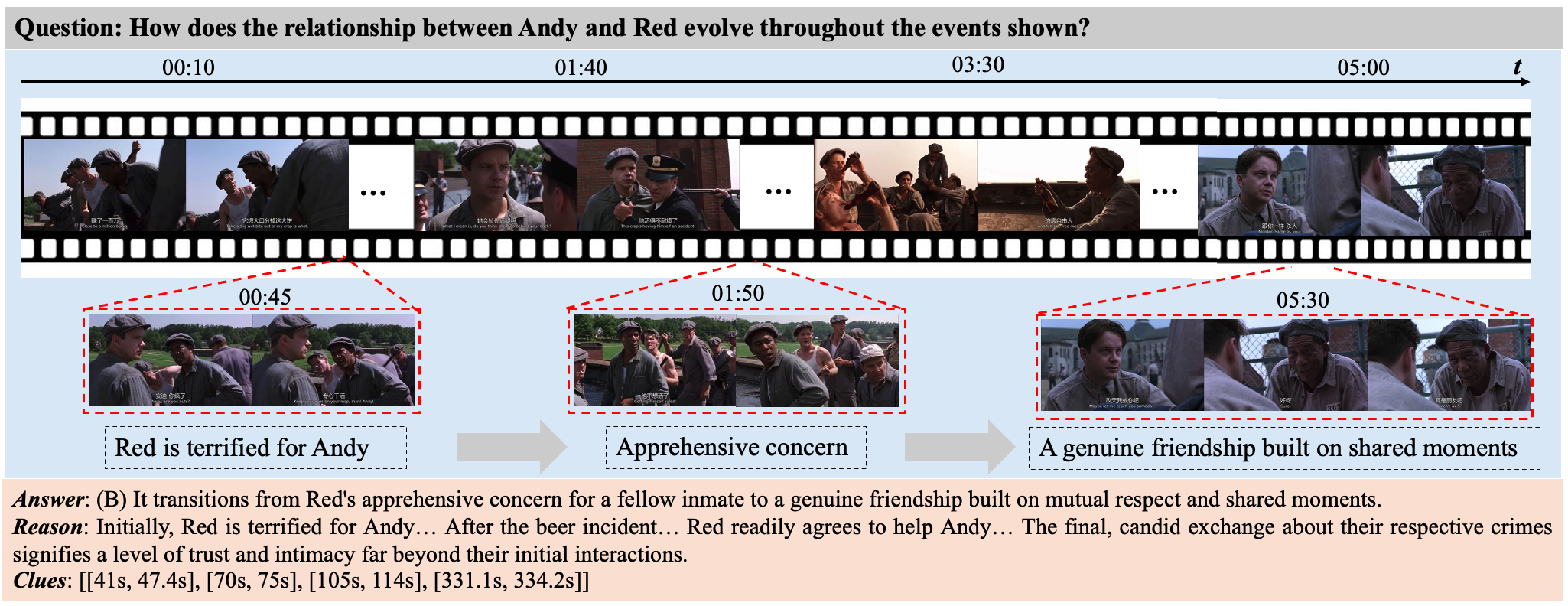}
        \caption{~A data example from the movie ``The Shawshank Redemption" in GLVC dataset.}
        \label{fig:dataset-example} 
    \end{subfigure}
    
    \caption{The overview of GLVC (Grounding Long Video Clues) dataset.}
    \label{fig:dataset}
    \vspace{-2em}
\end{figure*}

\subsection{Dataset Construction Pipeline}
\label{sec:dataset-construct}

Fig.~\ref{fig:dataset-construct} shows the process of dataset construction in detail. Our video data comes from IMDB Top $250$ high-rated movies. We first filtered out non-English movies and those before $1960s$. Then we crawled cast images from IMDB website for these movies, and leveraged face detection toolbox\footnote{\url{https://github.com/deepinsight/insightface}} and face re-identification model~\citep{chen2023beyond} to obtain character's images from movie videos. Due to fixed video lengths could disrupt the semantic integrity of plots, we divide the entire movie video into multiple segments based on the scenes. Specifically, we first collected corresponding OCR subtitles from opensubtitles website\footnote{\url{https://www.opensubtitles.com/}}, which includes subtitle data and timestamps. Then we leveraged the powerful reasoning ability of OpenAI o1 model to divide complete segments based on the semantics in the subtitles. These segments are manually checked, corrected, and filtered out inappropriate samples that are too long or too short. Finally, the character images, video segments, and corresponding subtitles are used as inputs. Based on the in-context learning method, we use Gemini 2.5 Pro model~\citep{comanici2025gemini} to annotate question-answer pairs, concrete reasoning and timestamps for the answer.

\subsection{Dataset Analysis}
\label{dataset:analysis}
The GLVC dataset includes $135$ movie videos, with a total duration of $284$ hours, covering $20$ movie genres.
There are a total of $10K$ training samples and $2K$ samples for evaluation. The video lengths vary between $2 \sim 10$ minutes.
Clues for each question are scattered throughout the entire video, which can more effectively evaluate the model's long video understanding ability.

The evaluation of model's capabilities includes two parts: 1) The semantic correctness of reasons predicted by the model. 2) The soft temporal IoU between timestamps of critical clues predicted by the model and ground truth.
Since the specific timestamps of clues could be a vague and approximate range, we introduce a ``tolerance" $\theta$. When the difference between the predicted timestamp of the model and the ground truth is less than $\theta$, the prediction of the model for this timestamp is considered correct. 
These three metrics can more comprehensively evaluate whether the model truly understands long videos.
% It can be formalized as:
% \begin{gather}
%     % b_i = \mathbf{1}_{a_i > 10} = \begin{cases}
%     % 1 & \text{if } a_i > 10 \\
%     % 0 & \text{otherwise}
%     % \end{cases}
%     P = \{ \mathbf{1}_{|P_t - G_t| \leq \theta}\}
% \end{gather}

% Correct if ∣tpredicted−tgt∣<shift\text{Correct if } |t_{\text{predicted}} - t_{\text{gt}}| < \text{shift}Correct if ∣tpredicted​−tgt​∣<shift

% Compared to related movie datasets~\citep{he2024storyteller,yue2023movie101}, our M3QA dataset strictly follows data structure division of real-world movies during shooting, which ensures the completeness of video semantics, thereby improving data quality. Additionally, the questions take into count multiple aspects related to characters, time and plots with different granularities, which can more comprehensively evaluate model's question-answering capabilities.
% We provide the prompt templates used in constructing process, detailed statistics and data examples in \textit{supplementary materials}.
% Tab.~\ref{tab_dataset_comp} compares our M3PU dataset with some related datasets. It can be seen that compared with id-aware~\citep{ji2024ida} and Movie-101~\citep{yue2023movie101}, our movie dataset has richer modality data. Compared with the fixed 3min duration of movie-story~\citep{he2024storyteller}, the flexible duration on the one hand ensures the integrity of the scene video content, which is conducive to improving data quality, and on the other hand, it can more easily apply and evaluate models under a wide range of durations.

\section{Experiments}
\label{sec:exp}

\subsection{Implement Details}
The video data is sampled at $1fps$ and each frame is resized to $224 \times 224$.  
% The dual audio encoders consist of encoder part of Whisper-Large-v2 model~\citep{radford2023robust} and fine-tuned BEATs~\citep{chen2022beats} encoder. The sample rate of raw waveform is converted to $16KHz$. We use a learnable token for a $0.33s$ audio window in window-level Q-former, which outputs 88 textual tokens for a 30-second audio. 
The parameters of LLM is initialized from Qwen2.5-VL-7B~\citep{bai2025qwen2}.
The trainable memory token embedding is initialized with \texttt{bos} token embedding of LLMs, and three matrices $W_{q,k,v}^m$ are initialized with the parameters of $W_{q,k,v}$ in LLMs. 
% The compression ratio $\alpha$ is selected from $4, 8, 16, 32$.
We employ a two stage training strategy, including warmup and long video training stages.

% During pre-training, we use $960$-hour LibriSpeech training set~\citep{panayotov2015librispeech} and $1000$-hour GigaSpeech M-set~\citep{chen2021gigaspeech} for speech recognition. The audio window-level Q-former is optimized for $1$ epoch with global batch size $256$ and learning rate $1e-4$.
% During fine-tuning, the memory-related trainable parameters are trained on our M3QA training set for $2$ epochs with global batch size $128$ and learning rate $1e-4$.
\textbf{Warmup}
We found that in the early stage of training, the model has not yet learned to make predictions based on memory tokens for long videos, resulting in high losses and gradients. To maintain training stability, we first warmup the memory-related trainable parameters. Specifically, we collected $8K$ pairs of video-caption data, uniformly sampling up to $60$ frames at $1fps$ for each video. Each video segment contains $32$ frames, with a fixed compression ratio of $32$. The learning rate is \num{1e-6}. This warmup stage allows the model to learn to understanding the entire video content using a limited number of memory tokens. 

\textbf{Long video training}
In this stage we collect videos from NExT-QA~\citep{xiao2021next}($32K$), LongVideo-Reason~\citep{chen2025scaling}($51K$), MovieChat~\citep{song2024moviechat}($1K$), Cinepile~\citep{rawal2024cinepile}($25K$) and GLVC dataset for long video question-answering training with a global batch size $64$ and learning rate \num{1e-4}.  The compression ratio is $16$. The video is uniformly sampled up to $512$ frames at $1fps$. Each video segment contains $32$ frames.
% The compression ratio is selected from $4, 8, 16, 32$. The video is uniformly sampled up to $512$ frames at $1fps$. Each video segment contains $32$ frames.

% Each video is sampled uniformly with 10 frames, and the size of each frame is $224 \times 224$. We extract the last layer patch-level representations for each frame. Following BEATs~\citep{chen2022beats}, we convert the sample rate of each raw waveform to $16KHz$ and extract 128-dimensional Mel-filter bank features with 25ms Povey window that shifts every 10ms as the acoustic feature. For the mask decoder, we use two \texttt{<MASK>} token groups corresponding to two scales of visual features from visual encoder, and we utilize the visual features from the 14th and second-to-last layers. Each group has three tokens. The number of query tokens in both audio Q-Former and visual Q-Former is 32. We use the LLaMA-2-7b-Chat model as our base model. The interaction-aware LoRA structure is employed in all linear layers with a rank of 8. The hyper-parameters $\lambda_{txt}$, $\lambda_{seg}$, $\lambda_{bce}$, $\lambda_{dice}$, and $\lambda_{ce}$ are set to $1.0, 0.5, 1.0, 0.5, 1.0$ respectively.
% More details in the \textit{supplementary materials}.

% \input{tabs/tab_main}

\subsection{Quantitative Results and Analysis}
\textbf{Evaluation benchmarks} We evaluate VideoDetective's ability on two types of video benchmarks: \textbf{1) Long video understanding benchmarks}, including Video-MME~\citep{fu2024video}, LongVideoBench~\citep{wu2024longvideobench}, MLVU~\citep{zhou2024mlvu}, VideoVista~\citep{li2024videovista} and Cinepile~\citep{rawal2024cinepile}. \textbf{2) Short video question answering benchmarks}, including MVBench~\citep{li2024mvbench} and NextQA~\citep{xiao2021next}.
These benchmarks mainly encompass question-answering tasks in long and short video scenarios.

\textbf{Compared models} The compared models mainly include three categories: 
\textbf{1) Closed-source models}: existing pioneering commercial models, which typically have long context lengths and powerful general video understanding capabilities, including Gemini 1.5 Pro~\citep{reid2024gemini}, GPT-4o~\citep{gpt4o} and Claude 3.5 Sonnet~\citep{Claude3}, \emph{etc.};
\textbf{2) Open-source general models}, which are usually trained on massive video data and possess general multi-task understanding capability, including mPLUG-Owl3~\citep{ye2024mplug}, MiniCPM-V 2.6~\citep{yao2024minicpm}, InternVL2~\citep{chen2024internvl}, LLaVA-Next-Video~\citep{zhang2024llavanext-video} and VideoLLaMA 2~\citep{cheng2024videollama}, \emph{etc.}.
\textbf{3) Open-source long video understanding models}, which are specifically designed for understanding long videos, including LLaMA-VID~\citep{li2024llama}, LongVA~\citep{zhang2024long}, LongLLaVA~\citep{wang2024longllava}, Video-XL~\citep{shu2025video} and LongVILA~\citep{chen2024longvila}, \emph{etc.}.

% To evaluate the performance of our method in long video question-answering tasks, we conducted evaluations on the M3QA test set and two other long video benchmarks: Cinepile~\citep{rawal2024cinepile} and Video-MME~\citep{fu2024video}. M3QA and Cinepile are minutes-long videos, and Video-MME consists of three types of videos: seconds-long short videos, minutes-long medium videos, and hours-long long videos. 

% The compared models mainly include two categories: 1) \textbf{Closed-source MLLMs}: existing powerful models, such as GPT-4o, GPT-4v and Gemini 1.5 Pro, \emph{etc.}; 2) \textbf{Open-source MLLMs}: including VideoLLaMA 2~\citep{cheng2024videollama} and video-SALMONN~\citep{sun2024video}, which are trained on large-scale audio-visual data and have strong short video understanding capabilities; LLaMA-VID~\citep{li2024llama} and Qwen2.5-VL~\citep{bai2025qwen2}, which can process long videos.

\textbf{Evaluation results across multiple video benchmarks.}
We present the performance of VideoDetective on popular long and short video benchmarks in Tab.~\ref{tab:main}.
Compared with closed-source advanced models with massive parameters and powerful general capabilities, there is still some performance gap on these benchmarks.
On the long video understanding benchmarks, VideoDetective surpasses most models, including the open-source general models with similar number of parameters (mPLUG-Owl3 and MiniCPM-V 2.6), as well as larger models (34B and 72B). 
Compared to other long video understanding models with same scale, VideoDetective also demonstrates excellent performance, achieving the best results on the Video MME, MLVU and VideoVista benchmarks, especially for medium and long videos in VideoMME, highlighting VideoDetective's strong capability in understanding long videos relying on seeking critical clues.
It is worth noting that on the LongVideoBench, VideoDetective lags behind other models. This is mainly because LongVideoBench features video-language interleved inputs up to an hour long. In our method, each video sub-segment is approximately $30 \sim 60$ seconds long, which would include a significant amount of language content, thus increasing the context length of each sub-segment, making it more challenging to seek clues.
VideoDetective also performs well on short video benchmarks. Specifically, it surpasses other open-source models on MVbench and achieves comparable performance on NextQA and the short video testing in VideoMME.

% This demonstrates VideoDetective's strong capability in understanding long videos, which relies on identifying key clues.

% To further evaluate the performance of our method on long video question answering tasks, we evaluate it on two long video benchmarks: Cinepile~\citep{rawal2024cinepile} and Video-MME~\citep{fu2024video}.
% Cinepile is designed for long-form video question answering and comprises questions convering six different aspects. The average video length in Cinepile is $160$ seconds.
% The Video-MME encompassing both short, medium, and long-term videos, ranging
% from $11$ seconds to $1$ hour. Tab.~\ref{tab_cinepile} and Tab.~\ref{tab_video_mme} show the evaluation results on these two benchmarks respectively. On Cinepile benchmark, our method achieves the best results among open-source models and comparable performance to GPT-4 Vision and GPT-4o ($55.36\%$ \emph{vs.} $55.35\%$ and $56.06\%$).
% On the Video-MME benchmark, it can be found that compared to models of the same size 7B, our method achieves the best results. Compared with the 34B LLaVA-NeXT-Video, we achieve the comparable performance on minutes-long medium videos ($49.9\%$ \emph{vs.} $50.1\%$) and superior results on hours-long long videos ($45.8\%$ \emph{vs.} $44.3\%$). While our method is only trained on minutes-long videos from M3QA dataset, it can easily generalize to hours-long scenarios, indicating it has great potential in ultra-long video understanding tasks.

\begin{table*}[t]\small
\centering

\vspace{-4mm}
\addtolength\tabcolsep{-2.4pt} 
\resizebox{0.95\linewidth}{!}{
\begin{tabular}{lc|cccc|cccc|c|c|c|c|c|c}
\toprule
\multicolumn{1}{c}{\multirow{2}{*}{\textbf{Model}}} & \multicolumn{1}{c|}{\multirow{2}{*}{\textbf{Size}}} & \multicolumn{4}{c|}{\textbf{VideoMME (w/o sub.)}} & \multicolumn{4}{c|}{\textbf{VideoMME (w/ sub.)}} & \multicolumn{1}{c|}{\multirow{2}{*}{\textbf{MLVU Test}}} & \multicolumn{1}{c|}{\multirow{2}{*}{\textbf{LVBench}}} &  \multicolumn{1}{c|}{\multirow{2}{*}{\textbf{VideoVista}}} & \multicolumn{1}{c|}{\multirow{2}{*}{\textbf{MVBench}}} & \multicolumn{1}{c|}{\multirow{2}{*}{\textbf{Cinepile}}} & \multicolumn{1}{c}{\multirow{2}{*}{\textbf{NextQA}}} \\ 

\multicolumn{1}{c}{} & \multicolumn{1}{c|}{}  & Short & Medium & Long & Avg. & Short & Medium & Long & Avg. & \multicolumn{1}{c|}{} & \multicolumn{1}{c|}{} & \multicolumn{1}{c|}{} & \multicolumn{1}{c|}{}  & \multicolumn{1}{c|}{} & \multicolumn{1}{c}{} \\

\midrule
\rowcolor{gray!15}\multicolumn{16}{c}{\textbf{Closed-source Pioneering MLLMs}} \\
% GPT-4V & - & 70.5 & 55.8 & 53.5 & 59.9 & & & & \\
Gemini-1.5-Pro~\citep{reid2024gemini} & - & 81.7 & 74.3 & 67.4 & 75.0 & 84.5 & 81.0  & 77.4 &  81.3 & - & 64.0 & - & 60.5 &  60.12 & - \\
GPT-4o~\citep{gpt4o} & - & 80.0 & 70.3 & 65.3 & 71.9 &  82.8 & 76.6 & 72.1 & 77.2 & 54.9 & 66.7  & 78.3 &  64.6 & 56.06 & - \\
Claude 3.5 Sonnet~\citep{Claude3} & - & 71.0 & 57.4 & 51.2 & 60.0 &  73.5 &  60.1 & 54.7 & 62.9  & - & - & - & - & - & - \\
\midrule

\rowcolor{gray!15}\multicolumn{16}{c}{\textbf{Open-source General MLLMs}} \\ 
% LOngViLA-R1 & 7B & 76.8 & 63.2 &  55.2 & 65.1 &  & & &  \\
mPLUG-Owl3~\citep{ye2024mplug} & 7B & 70.0 & 57.7 & 50.1 & 59.3 & 72.8 & 66.9 & 64.5 & 68.1 & 17.2 & 59.8 & - & -  & 38.27 & - \\
MiniCPM-V 2.6~\citep{yao2024minicpm} & 8B & 71.3 & 59.4 & 51.8 & 60.9 & 73.5 & 61.1 & 56.3 & 63.7  &  - & 54.9 & - & - & 46.91 & - \\
InternVL2~\cite{chen2024internvl} & 34B & 72.0 & 59.1 & 52.6 & 61.2 & 72.8 & 61.3 & 53.0 & 62.4 & 45.7 & 59.3 & - & 43.86 & - & - \\
LLaVA-Next-Video~\citep{zhang2024llavanext-video} & 34B & 61.7 & 50.1 & 44.3 & 52.0  & 65.1 & 52.2 & 47.2 & 54.9 & 50.5 & - & 56.7 & - &  - & - \\
VideoLLaMA 2~\citep{cheng2024videollama} & 72B & 69.8 & 59.9 & 57.6 & 62.4 & 72.0 & 63.0 & 59.0 & 64.7 &  45.6 & - & 60.5  &  34.1 & - & - \\

\rowcolor{gray!15}\multicolumn{16}{c}{\textbf{Open-source Long Video MLLMs}} \\ 
LLaMA-VID~\cite{li2024llama} & 7B & - & - & - & - & - & - &  - &  - & 17.2 & - & 56.9 &  41.4 & - & - \\
LongVA~\cite{zhang2024long} & 7B  & 61.1 & 50.4  & 46.2 & 52.6 & 61.6 & 53.6 & 47.6 & 54.3 & 41.1 & - & 67.4 & - & \underline{41.04} & - \\
LongLLava~\cite{wang2024longllava} & 7B & 61.9 & 51.4 & 45.4 & 52.9 & 66.2 & 54.7 & 50.3 & 57.1 &  - & -  & - & 49.1 & -  & - \\
Video-XL~\cite{shu2025video} & 7B & 64.0 & 53.2 & 49.2 & 55.5 & 67.4 & 60.7 & 54.9 & 61.0 & \underline{45.5} & 50.7  & \underline{70.6} & \underline{55.3} & - & - \\
LongVILA~\citep{chen2024longvila} & 7B & \textbf{69.0} & \underline{58.3}  & \underline{53.0} & \underline{60.1} & \textbf{72.9} & \underline{64.9} & \underline{57.4} & \underline{65.1} & - & \textbf{57.1} & - & - &  - & \textbf{80.7} \\
\midrule

\rowcolor{ModelGreen}\textbf{VideoDetective (Ours)} & 7B  & \underline{68.6} & \textbf{61.6} & \textbf{56.0} & \textbf{62.1}   & \underline{70.4} & \textbf{66.5} & \textbf{63.5} & \textbf{66.8}  & \textbf{45.8} & \underline{51.1} &  \textbf{74.3} &  \textbf{58.3} & \textbf{67.1}$^{*}$ & \underline{79.3} \\ 

\bottomrule
\end{tabular}
}
\vspace{-2mm}
\caption{Experimental results on mainstream short and long video benchmarks. * indicates   CinePile training set is included in training data, while results for other models are zero-shot evaluation.
} 
\vspace{-2mm}
\label{tab:main} 
\end{table*}

\subsection{Visual Needle-In-The-Haystack evaluation}
To explore VideoDetective's capability in handling long video inputs, we further conducted the Needle-In-The-Haystack evaluation~\citep{zhang2024long}. As shown in Fig.~\ref{fig:teaser-bc}(c), we primarily compared the baseline model Qwen2.5-VL and two long video understanding models, LongVA and LongLLava. Since the Qwen2.5-VL model lacks the capability to understand extremely long videos, it encounters memory overflow issues when the number of input frames reaches $800$. Although LongVA and LongLLava can handle longer video frames, such as $1.2K$ and $3K$, their performance significantly declines.
In contrast, VideoDetective, equipped with an efficient question-aware memory mechanism, can process up to $4k$ video frames and still achieve satisfactory results.

\begin{table*}[t]\small
\centering

\vspace{-1mm}
\addtolength\tabcolsep{-2.4pt} 
\resizebox{0.90\linewidth}{!}{
\begin{tabular}{c|cc|cc|cc|cc}
\toprule
\multicolumn{1}{c|}{\multirow{2}{*}{\textbf{Model}}}  & \multicolumn{2}{c|}{\textbf{Score}} & \multicolumn{2}{c}{\textbf{mIoU@5}} & \multicolumn{2}{c}{\textbf{mIoU@10}} & \multicolumn{2}{c}{\textbf{mIoU@15}} \\ 

\multicolumn{1}{c|}{} & \multicolumn{1}{c}{w/o sub.} & \multicolumn{1}{c|}{w/ sub.} & \multicolumn{1}{c}{w/o sub.} & \multicolumn{1}{c}{w/ sub.} & \multicolumn{1}{c}{w/o sub.} & \multicolumn{1}{c}{w/ sub.} & \multicolumn{1}{c}{w/o sub.} & \multicolumn{1}{c}{w/ sub.} \\

\midrule
 
% LongLLava~\cite{wang2024longllava} & 95.7 & 96.2 & 0.47 & 0.56 & 14.5 & 22.5 & 16.7 & 23.3 & 20.8 & 25.6 \\
% LongVA~\cite{zhang2024long} & 89.2 & 89.6  & 0.52 & 0.58 &  15.4 &  17.0  & 23.1 & 25.6 & 29.4 & 32.5 \\
% \midrule

% \rowcolor{ModelGreen}\textbf{VideoDetective (Ours)} & 93.7 & 94.4 & 0.62 & 0.69 & 23.5 & 26.6 & 36.2  & 37.3  & 42.0  & 44.9 \\ 

LongLLava~\cite{wang2024longllava} & 0.47 & 0.56 & 14.5 & 22.5 & 16.7 & 23.3 & 20.8 & 25.6 \\
LongVA~\cite{zhang2024long} & 0.52 & 0.58 &  15.4 &  17.0  & 23.1 & 25.6 & 29.4 & 32.5 \\
\midrule

\rowcolor{ModelGreen}\textbf{VideoDetective (Ours)} & 0.62 & 0.69 & 23.5 & 26.6 & 36.2  & 37.3  & 42.0  & 44.9 \\ 

\bottomrule
\end{tabular}
}
\vspace{-2mm}
\caption{Evaluation results on GLVC dataset. ``Score" represents the semantic similarity between the reasons predicted by model and actual reasons.
``mIoU@k" indicates tolerance $\theta = k$ seconds.
} 
\vspace{-2mm}
\label{tab:glvc} 
\end{table*}

% ``Score" The term "score" refers to the semantic similarity between the reasons predicted by the model and the actual reasons, as assessed using GPT.

% %%% warp table
% \begin{wraptable}{r}{0.45\textwidth}
% \centering
% \small
% \addtolength\tabcolsep{-2.4pt} 
% \begin{tabular}{c|cc|cc|cc|cc|cc}
% \toprule
% \multicolumn{1}{c|}{\multirow{2}{*}{\textbf{Model}}} & \multicolumn{2}{c|}{\textbf{Acc}} & \multicolumn{2}{c|}{\textbf{Score}} & \multicolumn{2}{c}{\textbf{mIoU@5}} & \multicolumn{2}{c}{\textbf{mIoU@10}} & \multicolumn{2}{c}{\textbf{mIoU@15}} \\ 

% \multicolumn{1}{c|}{} & \multicolumn{1}{c}{w/o sub.} & \multicolumn{1}{c|}{w/ sub.} & \multicolumn{1}{c}{w/o sub.} & \multicolumn{1}{c|}{w/ sub.} & \multicolumn{1}{c}{w/o sub.} & \multicolumn{1}{c}{w/ sub.} & \multicolumn{1}{c}{w/o sub.} & \multicolumn{1}{c}{w/ sub.} & \multicolumn{1}{c}{w/o sub.} & \multicolumn{1}{c}{w/ sub.} \\

% \midrule
 
% LongLLava & 95.7 & 96.2 & - & - & 14.5 & 22.5 & 16.7 & 23.3 & 20.8 & 25.6 & \\
% LongVA & 89.2 & 89.6  & - & - &  15.4 &  17.0  & 23.1 & 25.6 & 29.4 & 32.5 \\
% % LongVILA  & &  &  &  &  &   \\
% % Video-XL  & &  &  &  &  &   \\
% % LLaMA-VID  & &  &  &  &  &    \\
% \midrule

% \rowcolor{ModelGreen}\textbf{VideoDetective (Ours)} & 93.7 & 94.4 & - & - & 23.5 & 26.6 & 36.2  & 37.3  & 42.0  & 44.9 \\ 

% \bottomrule
% \end{tabular}
% \caption{Experimental results on GLVC dataset.} 
% \label{tab:glvc} 
% \end{wraptable}

\textbf{Experimental results on GLVC dataset}
Although the Visual Needle-In-The-Haystack evaluation can benchmark model's ability to accurately retrieve information from long context, the 
``needle" in this evaluation is merely a single image inserted at one position in an hour-long video.
% The "Needle" evaluation, which simply involves inserting a single image in an hour-long video.
To more effectively measure the model's capability to seek critical clues from long videos, we further evaluated the performance of long video models on the GLVC dataset. As shown in Tab.~\ref{tab:glvc}, it can be found that compared with LongVA and LongLLaVA models, Video Detective achieves higher GPT scores and mIoU at $\theta$ values of $5$s, $10$s, and $15$s, respectively. This indicates that, with the help of our question-aware memory mechanism, Video Detective can more effectively seek crucial clues. Additionally, subtitle data can also help enhance the model's long video understanding ability.

% \subsection{Memory Mechanism Visualization}
% As mentioned in Sec.~\ref{sec:intro} and shown in Fig.~\ref{fig:lora-attn}, in such long context and complex movie scenes, the model's attention to the entire video is usually indiscriminate, making it difficult to find crucial information.
% To verify the rationality of our memory mechanism, we visualized the attention scores of last layer in LLMs during inference, as shown in Fig.~\ref{fig:attn-vis}. For x-axis, the key sequence from left to right includes past memory tokens, current sub-segment tokens, and few current memory tokens. For y-axis, the past memory tokens are excluded. We can find that when the model recurrently processes sub-segments, it shows prominent attention scores for some critical key sequence, which is different from ordinary LoRA fine-tuning. This indicates that critical clues related to question is stored in these key sequences, which is consistent with the purpose of our proposed question-aware memory mechanism.

\subsection{Inference Efficiency}
We further evaluated the inference efficiency of our method compared to LoRA fine-tuning, LongVA and LongLLaVA, including inference time and GPU memory usage. As shown in Fig.~\ref{fig:teaser-bc} (Right), our method can process longer videos with comparable inference time but much lower memory usage.
Video Detective recurrently processes each seconds-long sub-segment, which does not lead to excessive memory usage while effectively seeking critical information related to question. More importantly, it can process an hour-long video at 1fps (3600 frames) while effectively seeking critical clues, which only requires 2 minutes and 37GB memory usage.

% \begin{figure}[t]
%     \vspace{-1em}
%   \centering
%   \begin{subfigure}[b]{0.23\textwidth}
%     \includegraphics[width=\textwidth]{iclr2026/figs/ablation_shot_duration.pdf}
%     \caption{}
%     \label{fig:ablation-shot-duration}
%   \end{subfigure}
%   \hfill % Optional: adjust the horizontal spacing between the subfigures
%   \begin{subfigure}[b]{0.23\textwidth}
%     \includegraphics[width=\textwidth]{figs/ablation_ratio.pdf}
%     \caption{}
%     \label{fig:ablation-compression-ratio}
%   \end{subfigure}
%   \vspace{-0.8em}
%   \caption{(a) The ablation results on the sub-segment duration.
%   (b) The ablation results on the compression ratio of memory tokens.}
%   \label{fig:ablation}
%   \vspace{-1.5em}
% \end{figure}

% \subsection{Visualize Results}

\subsection{Ablation Results}

\begin{wrapfigure}[12]{r}{0.46\textwidth}
    \centering
    \includegraphics[width=0.40\textwidth]{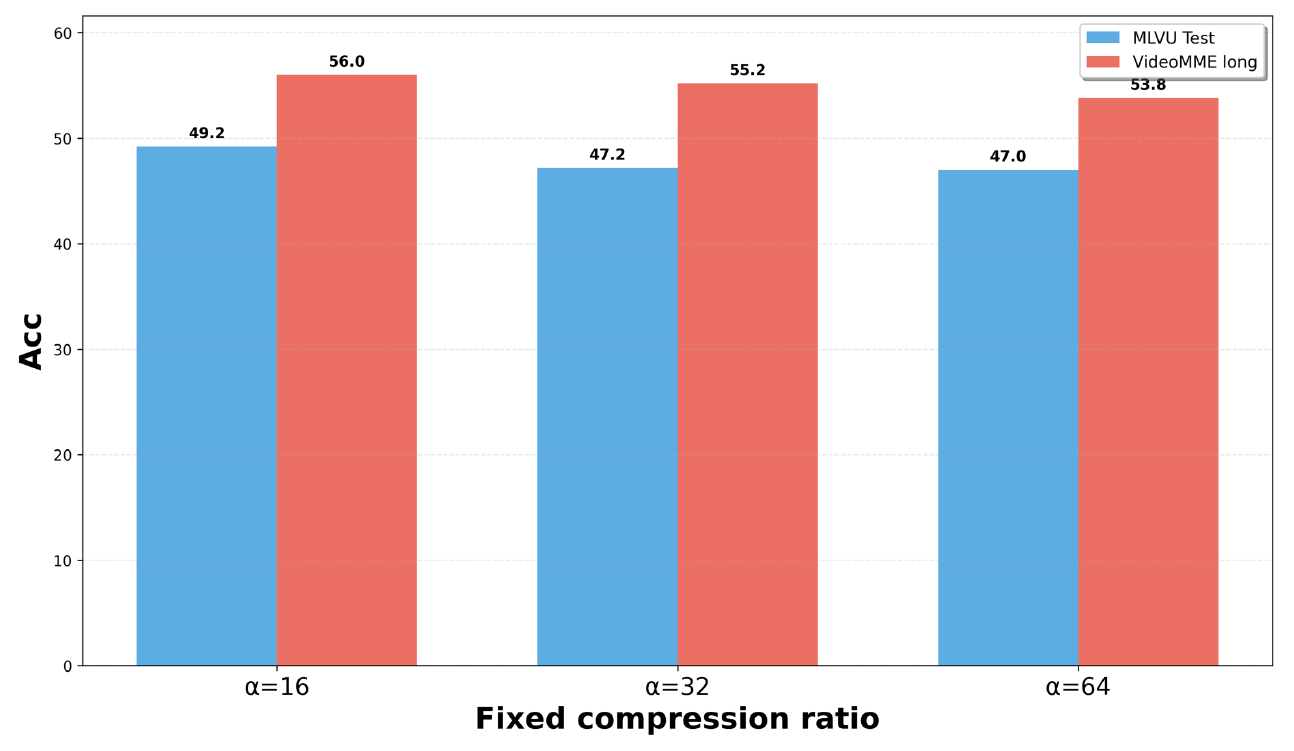}
    \caption{The impact of compression ratio $\alpha$.}
    \label{fig:ablation-compression}
\end{wrapfigure}

To validate effectiveness of our method, we conducted ablation experiments with different compression ratio; 
As shown in Fig.~\ref{fig:ablation-compression}, we evaluated model's performance on MLVU and VideoMME benchmarks with a fixed compression rate $16$, $32$ and $64$, respectively. Experimental results show that when compression rate is fixed, the model's performance gradually decreases as the compression ratio decreases. This is because a larger compression ratio will lead to fewer memory tokens, and thus less information can be sought.

%%% supp
% \textbf{Question-aware and history context.} Tab.~\ref{tab: ablation} shows the ablation results on question-aware compression of visual tokens and history context. ``\emph{w/o.} question-aware" means the memory tokens do not pay attention to question when processing each sub-segment. The experimental results show that they are an indispensable part of our question-aware memory mechanism. ``\emph{w/o.} history context" means previous memory tokens are not reused when processing each sub-segment, and only when predicting answers all the memory tokens are used. 

% \begin{figure}[t]
%      \centering
%      \includegraphics[width=0.46\textwidth]{iclr2026/figs/vis.png}
%      \vspace{-1em}
%      \caption{
%      The attention scores of sub-segments in the LLMs last layer during inference. (a) sub-segment 2. (b) sub-segment 3. (c) sub-segment 8. (d) sub-segment 9.
%      }
%      \label{fig:attn-vis}
%      \vspace{-1.25em}
% \end{figure}

% \input{tabs/tab_ablation_ratio}

\section{Conclusion}
\label{sec:conclussion}
In this work, we propose an efficient question-aware memory mechanism, enabling the model to recurrently seek critical clues from long videos. Experimental results on multiple long and short video benchmarks demonstrate the significant potential of our method in understanding long videos. Furthermore, to more effectively evaluate the model's ability to seek critical clues, we have constructed the GLVC dataset, which features concrete clues and allows for the quantitative assessment of the model's long video understanding capabilities.

\bibliography{iclr2026_conference}
\bibliographystyle{iclr2026_conference}

\appendix
\section{Appendix}

% \subsection{Prompt for dataset construction}

\subsection{dataset example}
\begin{figure*}[h]
    \centering
    \begin{subfigure}{\linewidth}
        \centering
        \includegraphics[width=0.95\linewidth]{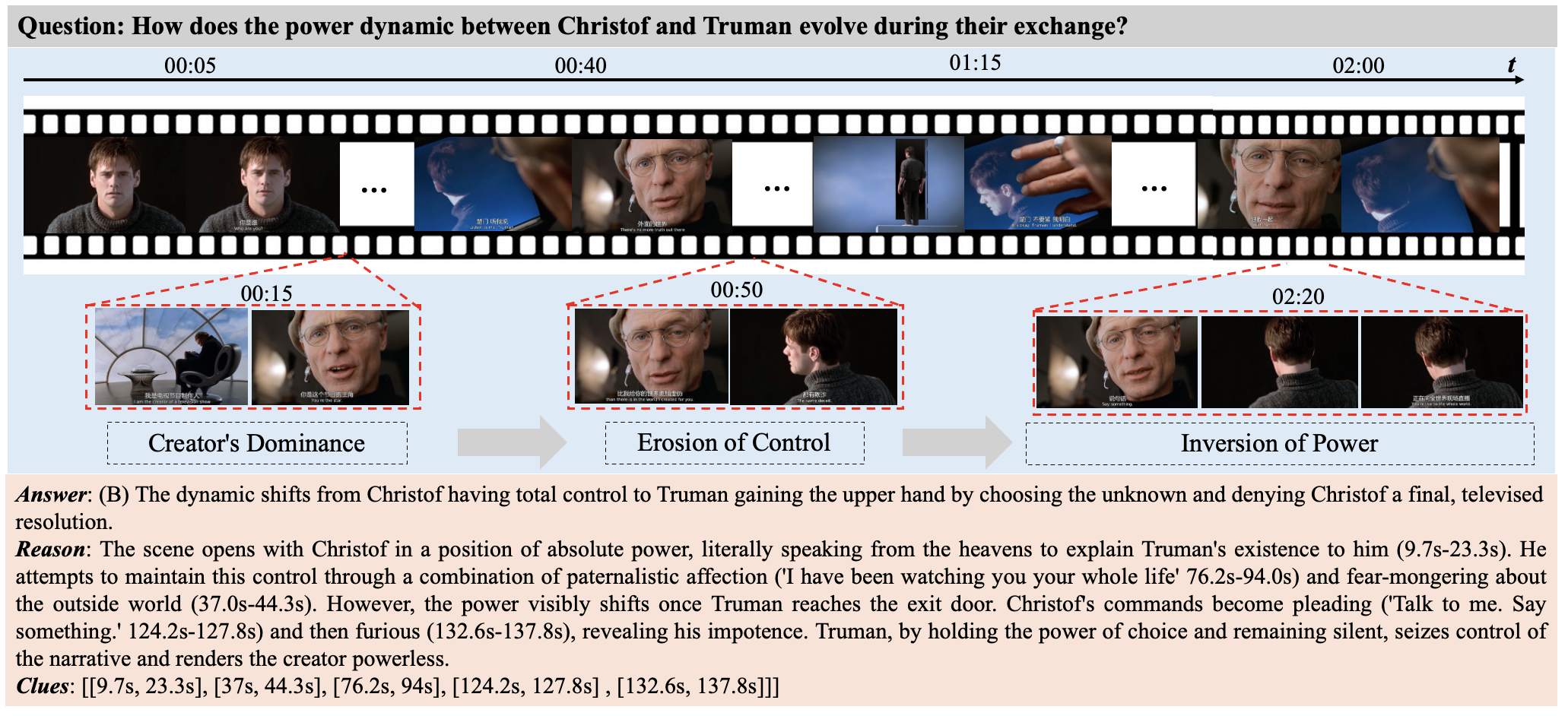}
        \caption{}
        \label{fig:example2}
    \end{subfigure}

    \begin{subfigure}{\linewidth}
        \centering
        \includegraphics[width=0.95\linewidth]{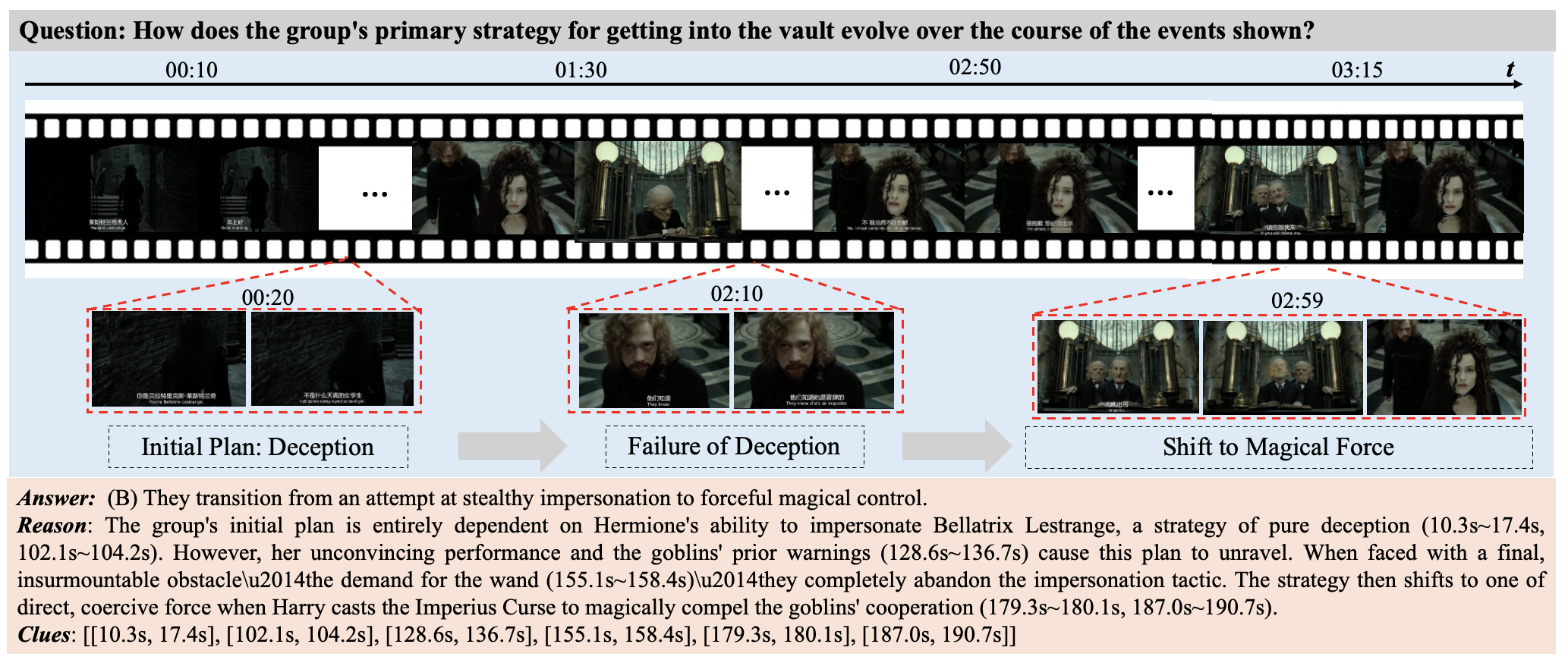}
        \caption{}
        \label{fig:example3}
    \end{subfigure}
    
    \caption{ Two data examples in our GLVC dataset comes from the movie ``The Truman Show" and ``Harry Potter and the Deathly Hallows".
    }
    \label{fig:data-example}
    \vspace{-2em}
\end{figure*}

To facilitate readers' understanding of our GLVC dataset, we have provided two additional data examples from the films "The Truman Show" and "Harry Potter and the Deathly Hallows." As shown in Fig.~\ref{fig:data-example}, answering the questions in these examples requires collecting sufficient clues from the entire video, rather than relying solely on a few video frames. These clues may be scattered across different corners of the video, allowing for a more comprehensive assessment of a model's ability to understand long videos.

% \subsection{visualize results}

% \subsection{inference efficiency}

\begin{wraptable}{r}{0.45\textwidth}
% \begin{table}[tbp]
\belowrulesep=0pt
\aboverulesep=0pt
\begin{center}
\caption{Ablation results on history context and question-aware compression of visual tokens.}
\vspace{-1em}
\label{tab: ablation}
% \abovetopsep=0pt
% \aboverulesep=0pt
% \belowrulesep=0pt
% \belowbottomsep=0pt
\scalebox{0.65}{
\begin{tabular}{c|c|c}
% \hline
% \bottomrule
\toprule
% \specialrule{1pt}{0pt}{0pt}
\textbf{Method} & \textbf{MLVU Test} & \textbf{NextQA} \\
\hline
\emph{w/o.} question-aware & 41.5 &  74.2 \\
\emph{w/o.} history context & 43.7 & 75.2  \\
\hline
\textbf{Video Detective (Ours)} & \textbf{45.8} & \textbf{79.3} \\
\bottomrule
\end{tabular}
}
% \hdashline[0.5pt/3pt]
\end{center}
\vspace{-2em}
% \end{table}
\end{wraptable}

\subsection{Ablation Results}
\textbf{Question-aware and history context.} Tab.~\ref{tab: ablation} shows the ablation results on question-aware compression of visual tokens and history context. ``\emph{w/o.} question-aware" means the memory tokens do not pay attention to question when processing each sub-segment. The experimental results show that they are an indispensable part of our question-aware memory mechanism. ``\emph{w/o.} history context" means previous memory tokens are not reused when processing each sub-segment, and only when predicting answers all the memory tokens are used.

% \subsection{attention score visualization process}

\subsection{LLM Usage}
We employed a large language model (LLM) solely for linguistic refnement of this manuscript,such as grammar correction, phrasing improvement, and style polishing. The LLM was not involvedin research design, data collection, model development, experiments,or analysis, All scientifccontributions, results, and conclusions are entirely the work of the authors.

\end{document}